\documentclass[lettersize,journal]{IEEEtran}
\pdfoutput=1
\usepackage{amsmath,amsfonts}
\usepackage{algorithmic}
\usepackage{algorithm}
\usepackage{array}
\usepackage[caption=false,font=normalsize,labelfont=sf,textfont=sf]{subfig}
\usepackage{textcomp}
\usepackage{stfloats}
\usepackage{url}
\usepackage{verbatim}
\usepackage{graphicx}
\usepackage{cite}
\usepackage{booktabs}
\usepackage{makecell}
\usepackage{multirow}
\usepackage{color}
\usepackage{float}
\begin{document}
	
	\title{ILNet: Low-level Matters for Salient Infrared Small Target Detection}
	
	\author{Haoqing Li, Jinfu Yang, Runshi Wang, Yifei Xu~\IEEEmembership{}
		\thanks{This work was supported in part by the National Natural Science Foundation of China under Grant no.61973009. (Corresponding author: Jinfu Yang.) 
			
		Haoqing Li, Runshi Wang, and Yifei Xu are with the Faculty of Information, Beijing University of Technology, Beijing, 100124, P. R. China. (e-mails: lihaoqing@emails.bjut.edu.cn; wangrunshi@emails.bjut.edu.cn; xuyifei@emails.bjut.edu.cn). 
			
		Jinfu Yang is with the Faculty of Information Technology, Beijing University of Technology, Beijing, 100124, China, also with the Beijing Key Laboratory of Computational Intelligence and Intelligent System, Beijing University of Technology, Beijing, 100124, P. R. China (e-mail: jfyang@bjut.edu.cn). }
		}
	
	\markboth{Journal of \LaTeX\ Class Files,~Vol.~14, No.~8, August~2022}%
	{Shell \MakeLowercase{\textit{et al.}}: A Sample Article Using IEEEtran.cls for IEEE Journals}
	
	\IEEEpubid{}
	
	\maketitle
	
	\begin{abstract}
	Infrared small target detection is a technique for finding small targets from infrared clutter background. Due to the dearth of high-level semantic information, small infrared target features are weakened in the deep layers of the CNN, which underachieves the CNN’s representation ability.
	
	To address the above problem, in this paper, we propose an infrared low-level network (ILNet) that considers infrared small targets as salient areas with little semantic information. Unlike other SOTA methods, ILNet pays greater attention to low-level information instead of treating them equally. A new lightweight feature fusion module, named Interactive Polarized Orthogonal Fusion module (IPOF), is proposed, which integrates more important low-level features from the shallow layers into the deep layers. A Dynamic One-Dimensional Aggregation layers (DODA) are inserted into the IPOF, to dynamically adjust the aggregation of low dimensional information according to the number of input channels. In addition, the idea of ensemble learning is used to design a Representative Block (RB) to dynamically allocate weights for shallow and deep layers. Experimental results on the challenging NUAA-SIRST (78.22\%nIoU and 1.33×10$^{-6}$Fa) and IRSTD-1K (68.91\%nIoU and 3.23×10$^{-6}$Fa) dataset demonstrate that the proposed ILNet can get better performances than other SOTA methods. Moreover, ILNet can obtain a greater improvement with the increasement of data volume. Training code are available at https://github.com/Li-Haoqing/ILNet.
	
	\end{abstract}
	
	
	\section{Introduction}
	\IEEEPARstart{A}{t} present, infrared small target detection (IRSTD) has been gradually applied to forest search and rescue, maritime rescue, fire prevention, and other fields \cite{bib1, bib2, bib3}. It is a technique to identify and locate the targets from a complex clutter background. As its imaging is not affected by light intensity and electromagnetic interference and has a certain penetration ability for clouds, it has some advantages over visible light detection \cite{bib4}. However, there are some characteristics on infrared small target detection:
	
	1) Small infrared targets tend to be smaller than 15 × 15 pixels compared to conventional visible targets.
	
	2) Low-level features such as color and texture are insufficient, while simultaneously the high-level semantic features of small infrared targets are often unrecognized.
	
	3) Infrared images possess inherent drawbacks, i.e. low signal-to-clutter ratio (SCR) and more noise than visible images. Consequently, small infrared targets are easy to submerge and difficult to recognize.
	
	These characteristics impose serious constraints on infrared small target detection and are prone to false alarm and miss detection.
	\begin{figure}[!t]
		\centering
		\includegraphics[width=3.8in]{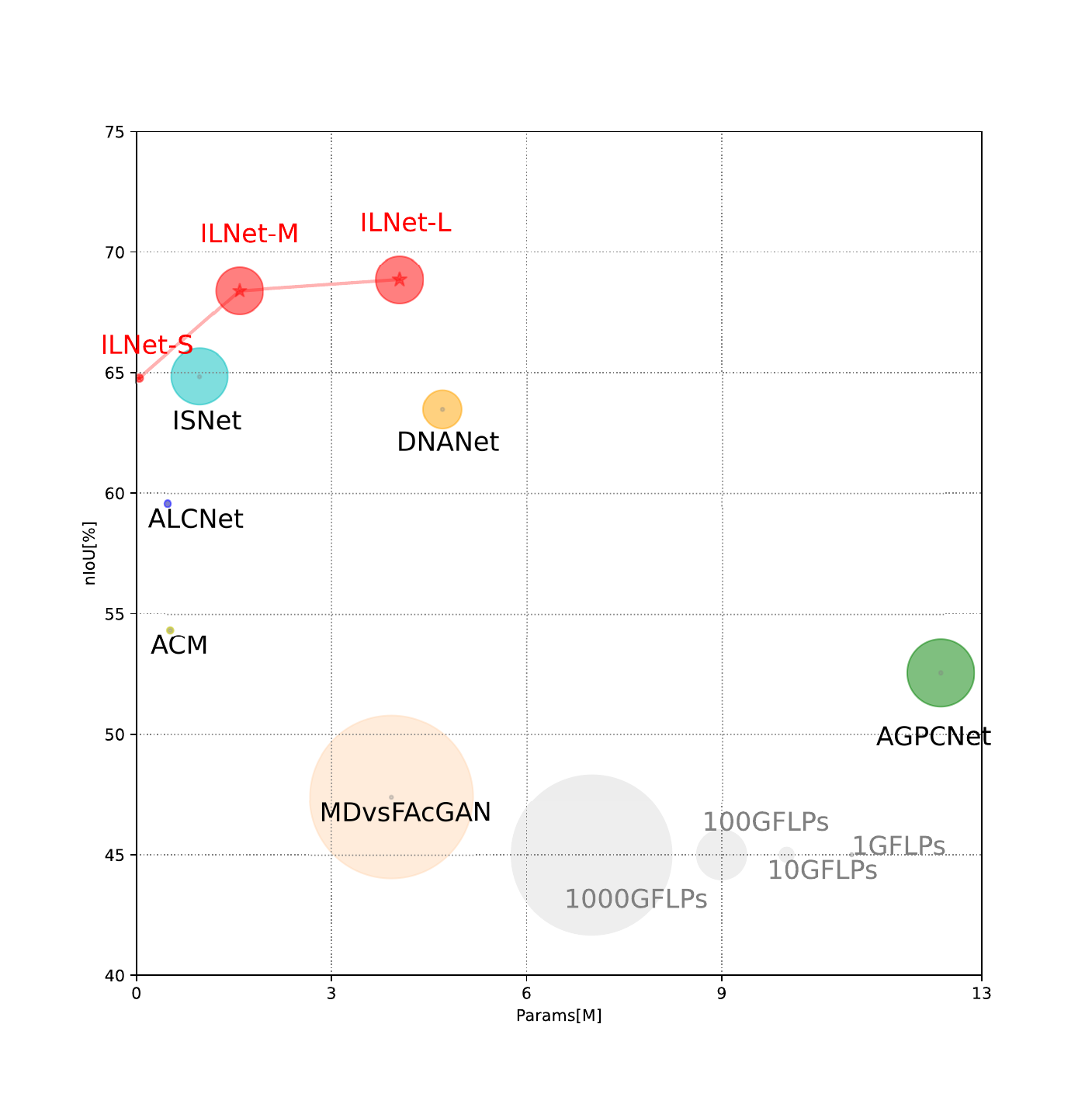}
		\caption{\textbf{nIoU \emph{vs.} size $\propto$ parameters, floating point operations.} Amount of parameters versus nIoU. The size of the blobs is proportional to the floating point operations; a legend is reported in the bottom right corner. Grey dots and stars highlight the centre of the blobs. Better near the upper left corner.
		}
		\label{Fig1}
	\end{figure}

	Infrared small target detection methods are divided into traditional methods and deep learning methods. Traditional methods include Filtering based \cite{bib5, bib6}, Local Contrast based \cite{bib7, bib8} and Local Rank based methods \cite{bib9, bib10, bib11, bib12, bib13}. The traditional model-driven methods are influenced seriously by numerous  hyper-parameters. Concurrently, their performance is relatively inferior, with high false alarm and miss detection.
	
	\begin{figure}[!t]
		\centering
		\includegraphics[width=3.6in]{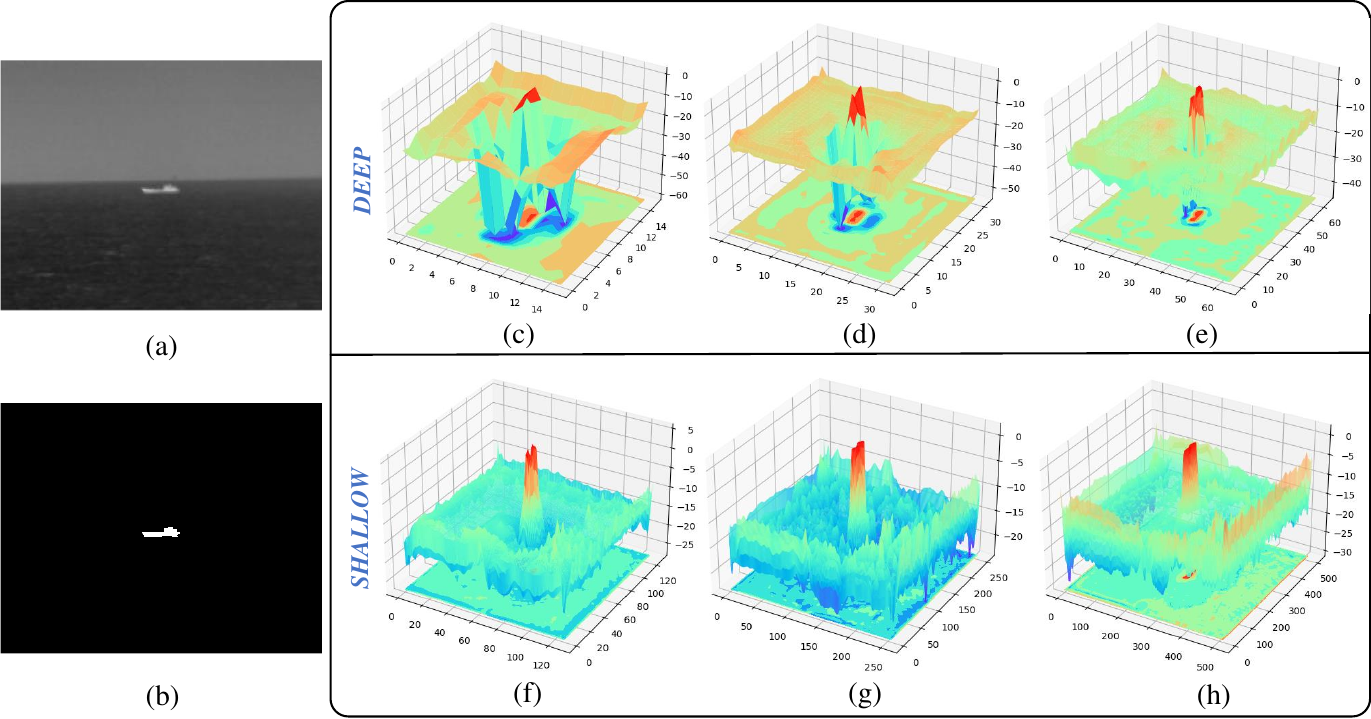}
		\caption{3D visualization of (a) input infrared image, (b) ground truth, (c)-(h) baseline decoder0-decoder5 output feature maps. Severe deep abatement exists within deep stages.}
		\label{Fig2}
	\end{figure}
	
	In recent years, the emergence of deep learning has triggered off a revolution in infrared small target detection. Deep learning methods consider complex infrared images as data, which are merely considered to be statistical samples or 0/1 in the computer. The infrared small target detection is considered as semantic segmentation without much complicated infrared knowledge. At the same time, the new methods can be utilized concisely after training, without a large number of hyper-parameters. For example, Dai et al. \cite{bib14, bib15} analyzed the discrepancies between model-driven and data-driven methods, and proposed ACMNet and ALCNet that can be trained end-to-end. Wang et al. \cite{bib16} used the paradigm of GAN \cite{bib17} to separate false alarm and miss detection. Zhang et al. \cite{bib18} improved the edge fitting ability by reconstruction of the target shape.
	
	The methods mentioned above are excellent with high detection probability and low false alarm. Typically, they regard the infrared small target detection as a semantic segmentation problem. By utilizing a multi-scale feature fusion module, high and low-level semantic information is fused for segmenting. However, unlike visible images, infrared small targets do not possess copious semantic information. They are rarely informative in the deep layers of the network, or even disappear completely, which is called \textit{deep abatement} in this paper, as shown in Fig.\ref{Fig2}. Ordinary feature fusion merely introduces shallow low-level information through a shortcut, but cannot completely restore the disappeared part.
	
	This paper innovatively reconsiders infrared small target detection from the perspective of salient object detection. Infrared small targets are considered as significant areas without special semantic information, while other areas are the background. Typically, the feature fusion module \cite{bib14,bib15,bib19,bib20} has a smaller internal resolution, which is required for computation and memory efficiency. But infrared small targets are tiny, lack of fine details and high-level features. Therefore, keeping high internal resolution at a reasonable cost is preferable for feature fusion. For this to happen, this paper proposes an interactive polarized orthogonal fusion module for bidirectional feature fusion. To preserve the potential loss of infrared small targets information by down-sampling, IPOF keeps an relatively high internal resolution. Consequently, more low-level information is included in the deep decoders. The feature fusion module is utilized in both deep and shallow stages of the network, but the feature dimensions of different stages are distinct. Typically, this dimension ranges from 8 to 1024, or even more. Stationary feature fusion modules have inadequate adaptability of distinct dimensions and thus tend to have redundant or insufficient representation ability. Dynamic one-dimensional aggregation layers are adjusted dynamically according to their input embedding dimension, to assemble channel and spatial information adaptively. Therefore, partial important information and details in the input features are retained adaptively to improve the detection performance. For the deep abatement, different up-sampling methods of the deep feature maps have significant impacts. As shown in Fig.\ref{Fig3}, all of these up-sampling methods are unable to recover small targets (contour, intensity, etc.) well. To alleviate the problem and aggregate comprehensive information, a representative block is proposed, which enhances the up-sampled feature map and dynamically assigns weights based on the contributions of stages.

	• This study reconsiders infrared small target detection from the perspective of salient object detection, and proposes a infrared small target detection network ILNet.
	
	• An interactive polarized orthogonal fusion module (IPOF) for bidirectional feature fusion is designed, which improves the features fusion through a channel and spatial high-resolution interaction.
	
	• Dynamic one-dimensional aggregation layers (DODA) are proposed and incorporated into the IPOF, to adaptively aggregate features with widespread resolution.
	
	• A representative block (RB) is proposed to distinguish the importance of high-level and low-level information, improve the deep abatement, and dynamically fuse global features.

	\section{Related Work}
	In this section, we provide a brief overview of the main works related to infrared small target detection, feature fusion and salient object detection.
	
	\subsection{Infrared Small Target Detection}
	Traditional infrared small target detection algorithms such as max median \cite{bib5}, IPI \cite{bib13}, and RIPT \cite{bib10}, were formerly the most efficient algorithms. However, with the introduction of deep learning, the light of traditional methods is gradually fading. Dai et al. \cite{bib14} introduced the concept of data-driven methods earlier, and proposed an Asymmetric Contextual Modulation (ACM) approach for infrared small target detection. Zhao et al. \cite{bib21} aimed at the extreme imbalance between foreground and background and imposed a semantic constraint. Wang et al. \cite{bib16} decoupled the minimization of false alarm and miss detection, by the adversarial learning paradigm, and considered the role of local and global information. Zhao et al. \cite{bib22} regarded small targets as noise, detection as an image-to-image conversion, and introduced L2 loss to improve positioning accuracy. Li et al. \cite{bib19} proposed a dense nested attention network and a channel-spatial attention module for adaptive feature fusion and enhancement, to incorporate and exploit contextual information. Zhang et al. \cite{bib18} innovatively emphasized that shape matters, and incorporated shape reconstruction into infrared small target detection. Wu et al. \cite{bib23} introduced a \textit{U-Net in U-Net}
	framework, enabling the multi-level and
	multi-scale representation learning of objects. Lv et al. \cite{bib24} proposed a deep network framework
	that covers feature enhancement, interaction, and comparison. Ren et al. \cite{bib25} proposed a Region Super-Resolution Generative Adversarial Network (RSRGAN) to solve the super-resolution applied problems. Although these methods have tremendously reduced the false alarm and miss detection, the infrared small target does not possess rich high-level semantic information. Our study found that Taylor finite difference(TFD)-inspired \cite{bib18} can be effective, essentially because low-level information is more instructive, which deserves more attention.
	\begin{figure}[!t]
		\centering
		\includegraphics[width=3.6in]{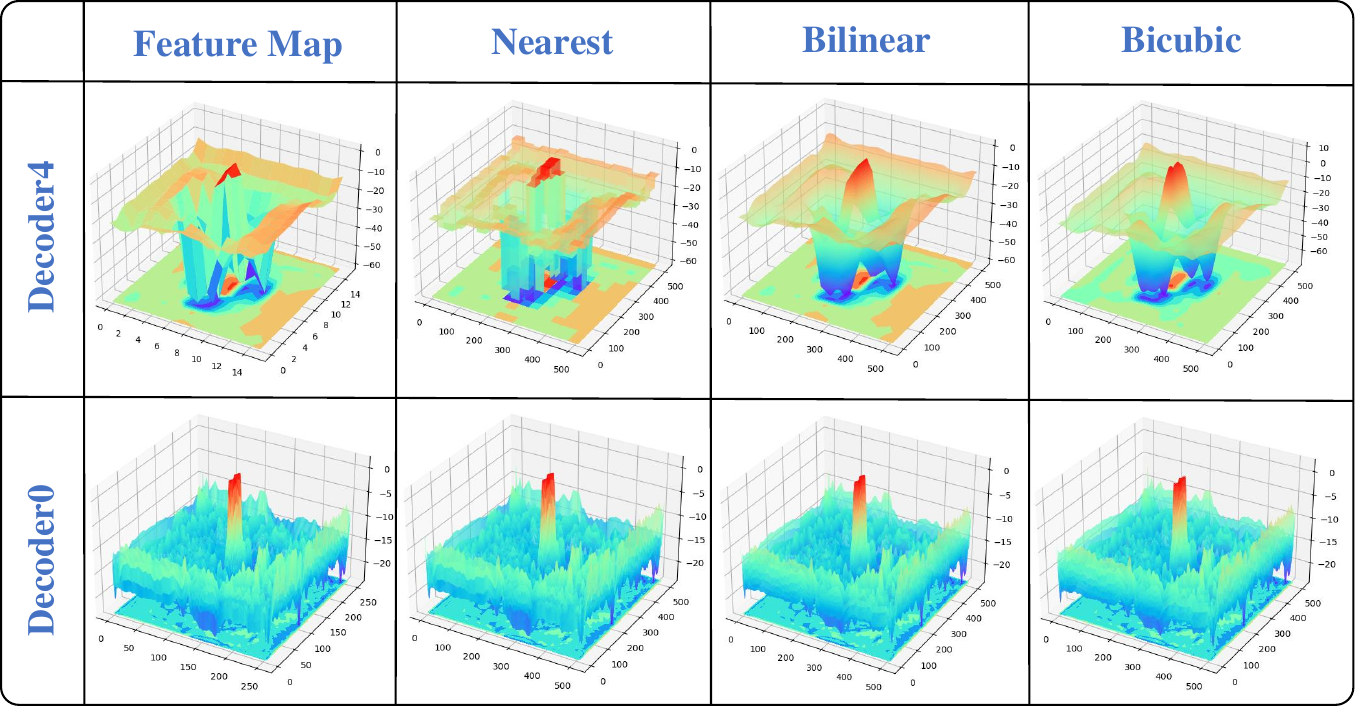}
		\caption{Different effects of up-sampling methods on deep and shallow stages of the baseline. These methods have little effect on the low-level features but a greater effect on the high-level.}
		\label{Fig3}
	\end{figure}
	\subsection{Feature Fusion}
	Deep learning methods often use encoding-decoding paradigms \cite{bib26, bib27, bib28} or feature pyramid paradigms \cite{bib29, bib30}. The methods above require feature fusion to combine low-level detail features and high-level semantic features. Intricate attention modules need to be designed during this process \cite{bib14, bib19}. For example, DNA-Net \cite{bib19} used dense nested attention among multiple layers to fuse and enhance features. Dai et al. \cite{bib14} proposed an Asymmetric Contextual Modulation (ACM) to fuse features of each scale, and designed a feature map cyclic shift scheme and bottom-up attentional modulation (BLAM) to retain the small target features. ISNet \cite{bib18} designed a two-orientation attention aggregation that facilitates shape information from low-level features, leading to a refinement of high-level features. Besides, this refinement should not be confined to shape, but take a broad view of all valuable low-level information.
	
	\subsection{Salient Object Detection}
	Salient object detection (SOD) focuses on the most salient areas and segments them at the pixel level. Infrared small targets are generally areas of high intensity in clutter background. They are visually salient without specific semantic information. Therefore, we consider them to be salient and small areas in infrared images. The foremost methods of SOD include multi-level deep feature integration methods \cite{bib31, bib32, bib33, bib34, bib35} and multi-scale feature extraction methods \cite{bib36, bib37, bib38}. In this paper, several modules are proposed to make the baseline suitable for small infrared target detection, and the performance surpasses the previous methods.
	
	\begin{figure*}[!t]
		\centering
		\setlength{\abovecaptionskip}{0.cm}
		\includegraphics[width=7.2in]{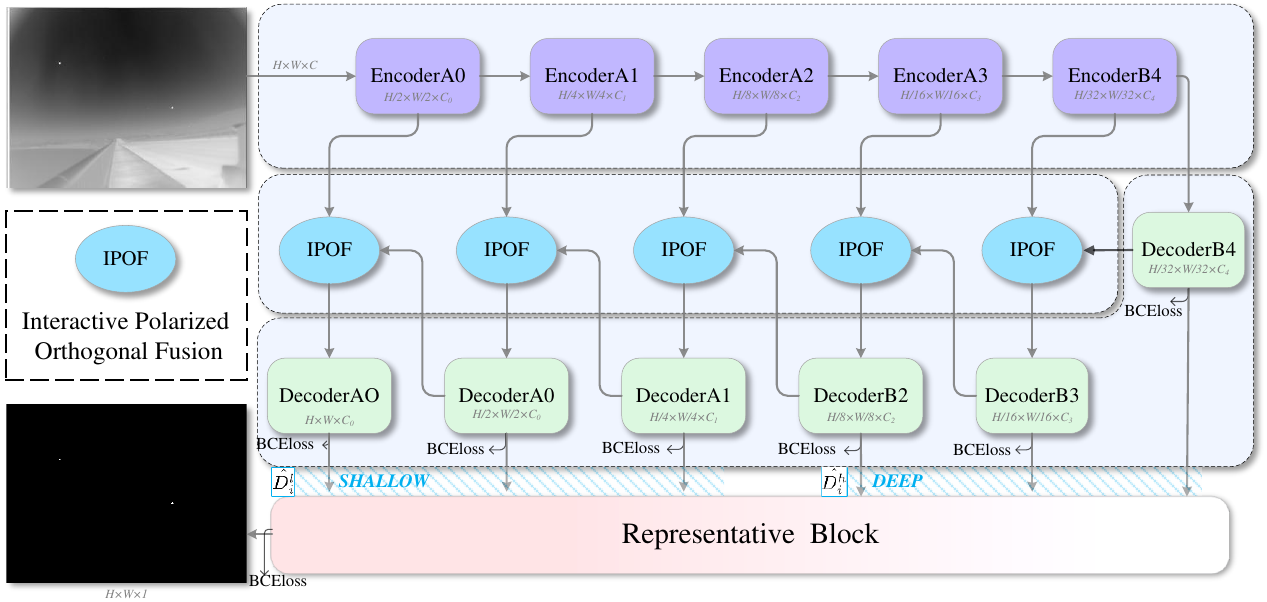}
		\caption{Overview of the proposed ILNet.}
		\label{Fig4}
		\vspace{-0.4cm}
	\end{figure*}

\begin{table}[t]
	\centering
	\caption{Number of the channels with different configurations. (Number of input, intermediate, output channels $C_i$).}
	\label{Li.t1}
	\renewcommand\arraystretch{1.5}
	\begin{tabular}{c|ccc}
		\Xhline{1.3pt}
		Stage&ILNet-L&ILNet-M&ILNet-S\\ 
		\Xhline{1pt}
		Encoder0&(3, 16, 64)&(3, 16, 64)&(3, 4, 8)\\
		Encoder1&(64, 16, 64)&(64, 16, 64)&(8, 4, 8)\\
		Encoder2&(64, 32, 64)&(64, 16, 64)&(8, 4, 8)\\
		Encoder3&(64, 32, 128)&(64, 16, 64)&(8, 4, 8)\\
		Encoder4&(128, 32, 128)&(64, 16, 64)&(8, 4, 8)\\
		Dncoder4&(128, 64, 128)&(64, 16, 64)&(8, 4, 8)\\
		Dncoder3&(256, 64, 128)&(128, 16, 64)&(16, 4, 8)\\
		Dncoder2&(256, 32, 64)&(128, 16, 64)&(16, 4, 8)\\
		Dncoder1&(128, 32, 64)&(128, 16, 64)&(16, 4, 8)\\
		Dncoder0&(128, 16, 64)&(128, 16, 64)&(16, 4, 8)\\
		DncoderO&(128, 16, 64)&(128, 16, 64)&(16, 4, 8)\\
		\Xhline{1.3pt}
	\end{tabular} 
\end{table}

	\section{Methodology}
	In this section, we first introduce the overall architecture of the ILNet (Section A). Then, we present the details of the IPOF module (Section B) and the DODA layers (Section C), followed by the RB block in Section D.
	\subsection{Overall Architecture}
	As shown in Fig.\ref{Fig4}, our ILNet is a U-shape encoder-decoder architecture network. Input goes through the encoders and decoders (i.e. ReSidual U-block) \cite{bib35}. First, the input infrared images are provided to the cascade encoder to extract the feature maps of different levels. Then, the feature maps are decoded gradually to the input dimensions through cascade decoders. At the same time, the outputs of the adjacent encoder and decoder are fused through the IPOF module for bidirectional feature fusion. Finally, the outputs of each decoder are gathered and fed into the RB block. Global information is aggregated discriminately to get the final output. 
	
	Section B details our feature fusion method, the interactive polarized orthogonal fusion module (IPOF), which contains DODA layers. Section C presents the dynamic one-dimensional aggregation layers (DODA) for adaptive aggregation of the channel and spatial information. In section D, we introduce the representative block (RB), which integrates the outputs of the decoder stages globally and dynamically.
	
	\subsection{Interactive Polarized Orthogonal Fusion Module}
	To incorporate more encoder low-level features into the decoder, the IPOF module is used to fuse the features of the same stages.
	
	The outputs of the $i^{th}(i = 0, 1, 2, ..., a)$ encoder and decoder stages are fetched, represented by $\mathbf{E}_i$ and $\mathbf{D}_i\in \mathbb{R}^{[N, C_i, H, W]}$ , as the input of the IPOF module. Where $N, C, H$, and $W$ denote the batchsize, number of channels, height and width of inputs, respectively. Spatial attention is separated from channel attention to balance FLOPs and fused information through dimension collapse \cite{bib39}.
	
\begin{figure*}[!t]
	\centering
	\setlength{\abovecaptionskip}{0.cm}
	\includegraphics[width=7.2in]{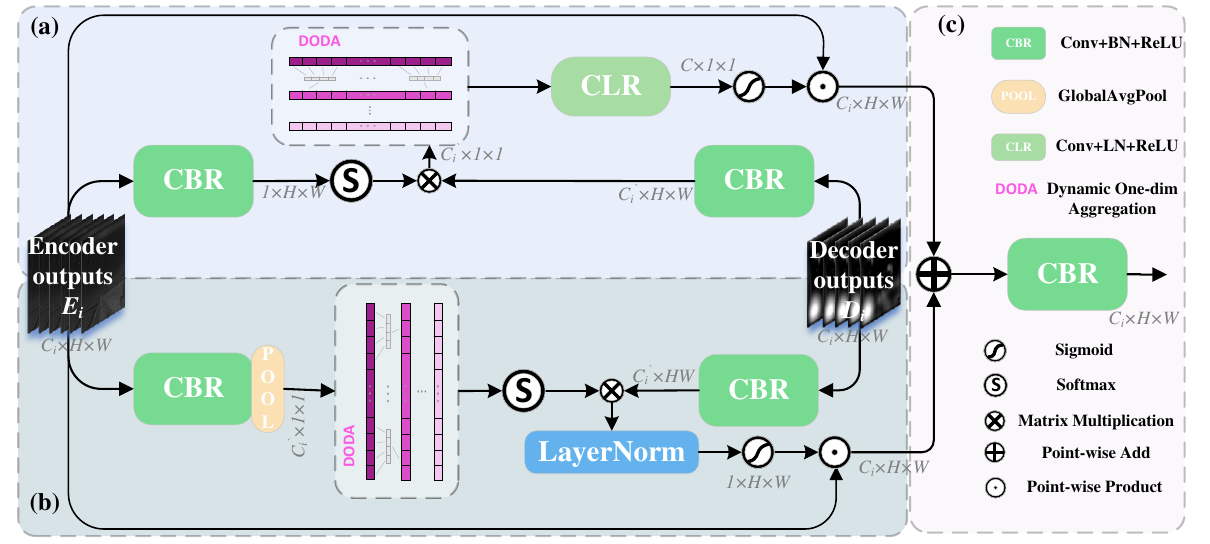}
	\caption{Interactive polarized orthogonal fusion module(IPOF). (a) Channel attention. (b). Spatial attention. (c) Aggregate and output.}
	\label{Fig5}
	\vspace{-0.4cm}
	\end{figure*}

	Spatial attention is shown in Fig.\ref{Fig5}(a). This part compresses the spatial dimension to 1 and calculates the attention map without over-sampling the channel dimension. The intermediate channel dimension $C_i^{\prime}$ is set to no less than $C_i/2$ generally, which retains enough dimension information. Formally, the spatial attention is instantiated as follows:
	\begin{equation}
		\label{1}
		\mathbf{A}_i^{S}(\mathbf{E}_i, \mathbf{D}_i)=\sigma(\varsigma_2(DODA[\varsigma_1(\mathbf{D}_i)\cdot S(\varsigma_1(\mathbf{E}_i))]))\odot \mathbf{E}_i,
	\end{equation}
	where $\sigma$ and $S$ denote the Sigmoid and Softmax function, respectively. $\varsigma_1$ is standard 1 × 1 convolution, batch norm, and ReLU layers. $\varsigma_2$ and $\varsigma_1$ are similar, except that batch norm is replaced by layer norm. $DODA$ is the proposed dynamic one-dimensional aggregation layer. $\odot$ represents the element-wise multiplication operator.
	
	Channel attention is shown in Fig.\ref{Fig5}(b). As with the spatial part above, this part collapses the input spatail dimension to 1 and preserves channel information as much as possible. Formally, the channel attention is instantiated as follows:
	\begin{equation}
		\label{2}
		\mathbf{A}_i^{C}(\mathbf{E}_i, \mathbf{D}_i)=\sigma(LN[S(DODA[P_{GA}(\varsigma_1(\mathbf{E}_i))])\cdot\varsigma_1(\mathbf{D}_i)])\odot \mathbf{D}_i,
	\end{equation}
	where $LN$ denotes the Layernorm, $P_{GA}$ is the golbal average pooling.
	
	In the output aggregation (Fig.\ref{Fig5}(c)), the parallel $\mathbf{A}^S_i(.)$ and $\mathbf{A}^C_i(.)$ are element-wise added, and then aggregated through a standard convolution layer. The final output can be formulated as:
	\begin{equation}
	\label{3}
	IPOF_i(\mathbf{E}_i,\mathbf{D}_i)=\varsigma_1(\mathbf{A}_i^C(\mathbf{E}_i,\mathbf{D}_i)+\mathbf{A}_i^S (\mathbf{E}_i,\mathbf{D}_i)).
	\end{equation}

	\begin{figure}[!t]
	\centering
	\includegraphics[width=3.4in]{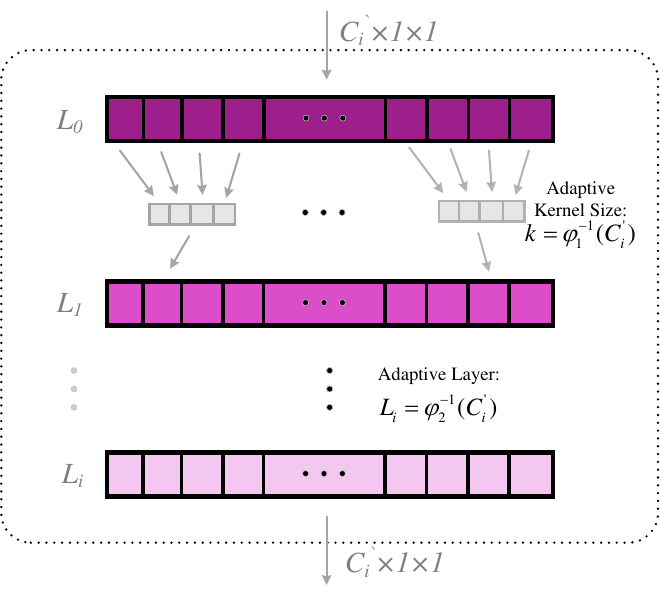}
	\caption{Dynamic One-dimensional Aggregation Layers.}
	\label{Fig6}
	\end{figure}

	\subsection{Dynamic One-dimensional Aggregation Layers}
	The dynamic one-dimensional aggregation layers is embedded in the IPOF module, as shown in Fig.\ref{Fig6}, for capturing local cross-channel interactions appropriately. Considering that there is a mapping $\varphi_1$ between convolution kernel $k$ and the number of input channels $C_i^\prime$ \cite{bib40}:
	\begin{equation}
		\label{4}
		C_i^\prime=\varphi_1(k)=2^{2k-1}.
	\end{equation}
	Since the $C_i^\prime(i=0,1,2,\dots,a)$ varies tremendously with $i$, the required number of 1D convolution layers $L_i$ also needs to be adjusted adaptively. There should be mapping $\varphi_2$ between $L_i$ and $C_i^\prime$:
	\begin{equation}
		\label{5}
		C_i^\prime=\varphi_2(L_i).
	\end{equation}
		
	\begin{figure*}[!t]
		\centering
		\includegraphics[width=7.2in]{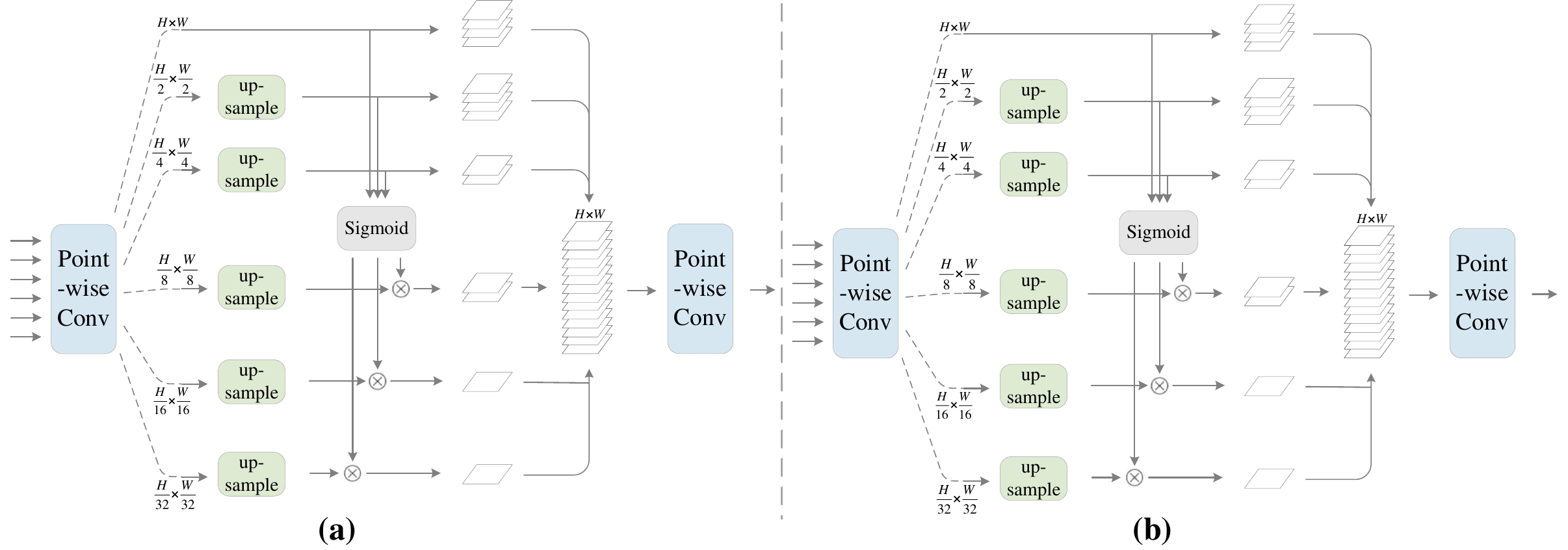}
		\caption{3D Visualization of RB Working Process.}
		\label{Fig7}
	\end{figure*}

	\begin{figure*}[!t]
	\centering
	\includegraphics[width=7in]{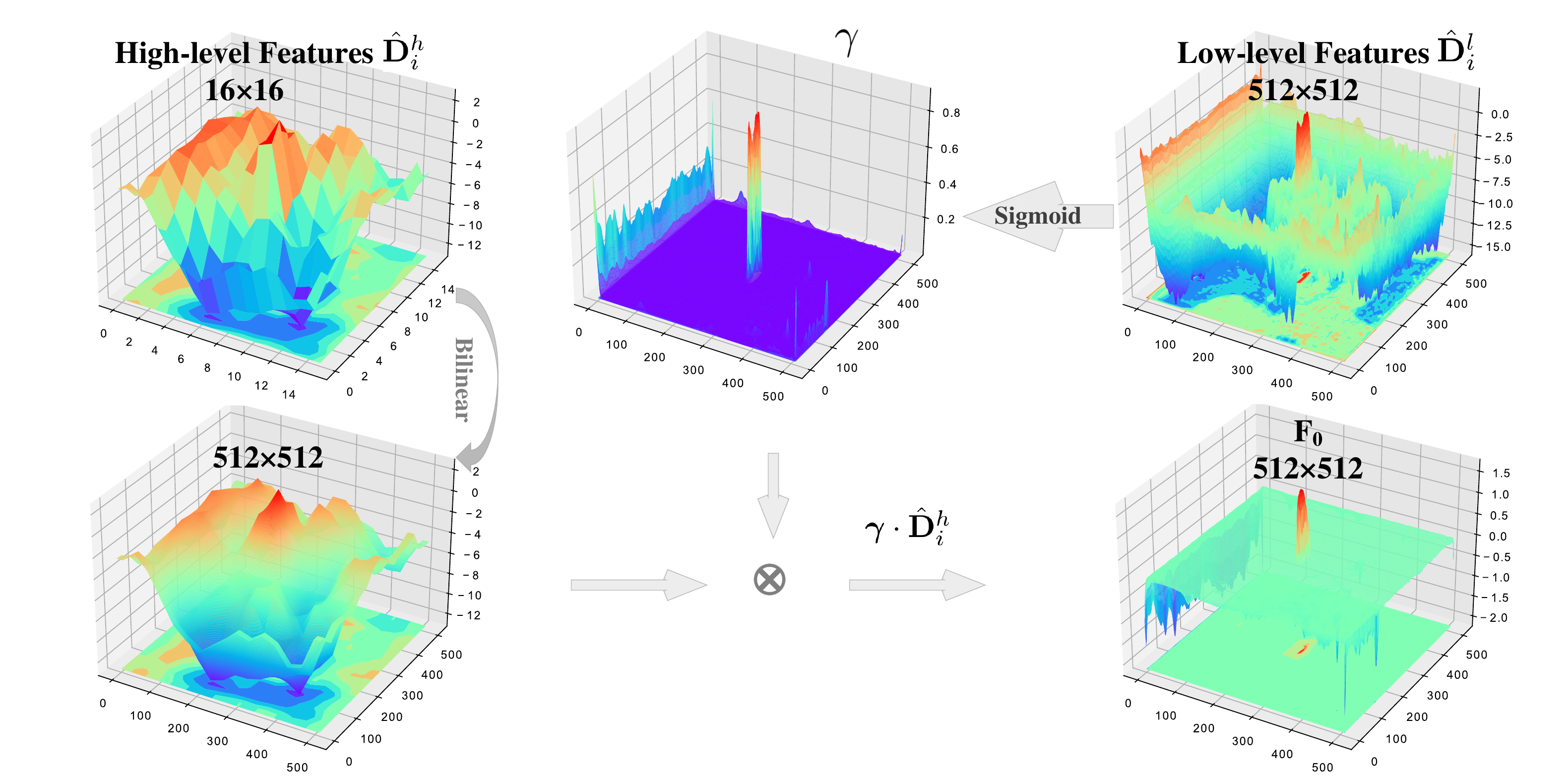}
	\caption{3D Visualization of RB Working Process.}
	\label{Fig8}
	\end{figure*}

	According to the practice, $C_i^\prime$ and $k$, $L_i$ present a certain degree of positive correlation frequently. Nonlinearity is introduced to eliminate its linear representation:
	\begin{equation}
		\label{6}
		C_i^\prime=\varphi_2(L_i)=2^{2n(L_i-1)+b},n,b\in \mathbb N^+.
	\end{equation}
	where $n$ and $b$ is hyper-parameters. Respectively, they are set to 2 throughout all the experiments.
	
	The number of 1D convolution layers $L_i$ is formulated as a non-linear mapping $\varphi_2:C_i^\prime\to L_i$, i.e., 
	\begin{equation}
		\label{7}
		L_i = 
			\lceil1-\frac{b}{2n}+\frac{\log_2\sqrt{C_i^\prime}}{n}\rceil.
	\end{equation}
	where $\lceil.\rceil$ is round up. Round down serves as choosing if a lightweight module is needed.

	\subsection{Representative Block}
	To improve the deep abatement after up-sampling in deep stages, and globally fuse the edge outputs of distinct decoders, the representative block is designed as shown in Fig.\ref{Fig7}. RB is more suitable for a small network with fewer parameters and inferior representation ability. Typically, edge outputs of the second half decoder stages, i.e. $\hat{\mathbf{D}}_i^l(i=\frac{a+1}{2},\dots,a)$ are considered as shallow outputs, where fuse more low-level features. Meanwhile, the first half outputs of the decoder stages, i.e. $\hat{\mathbf{D}}_i^h(i=0, 1,\dots,\frac{a-1}{2})$ are considered as the deep outputs, where a serious deep abatement exists, as is shown in Fig.\ref{Fig2}. The enhancement for up-sampled features needs to be implemented.
	
	$\hat{\mathbf{D}}_i^l(i=\frac{a+1}{2},\dots,a)$ is used to enhance $\hat{\mathbf{D}}_i^h(i=0, 1,\dots,\\\frac{a-1}{2})$. The low-level information in the shallow outputs is introduced into the deep stages, which enhances the low-dimensional features of the deep outputs. Moreover, the recovery ability of the small infrared targets after up-sampling is improved. This can be viewed as a special attention mechanism without many FLOPs. The shallow outputs $\hat{\mathbf{D}}_i^l(i=\frac{a+1}{2},\dots,a)\in \mathbb{R}^{[N, C_i, \frac{H}{d}, \frac{W}{d}]}$, are up-sampled to the output size directly,i.e. $\mathbf{F}_i\in \mathbb{R}^{[N, C_i, H, W]}(i=\frac{a+1}{2},\dots,a)$:

	\begin{equation}
		\label{8}
		\boldsymbol{\gamma}=Sigmoid(Conv(Cat[\hat{\mathbf{D}}_\frac{a+1}{2}^l, \dots, \hat{\mathbf{D}}_a^l])),
	\end{equation}
	\begin{equation}
		\label{9}
		\mathbf{F}_i = \begin{cases}
			\boldsymbol{\gamma}\cdot\hat{\mathbf{D}}_i^h,&{\text{if}}\ i>\frac{a}{2} \\ 
			{\hat{\mathbf{D}}_i^l,}&{\text{otherwise,}} 
		\end{cases}
	\end{equation}

	where $d$ is the down-sampling rate, $\boldsymbol{\gamma}\in\mathbb{R}^{[N, 1, H, W]}$ is the enhancement factor, $Conv(.)$ is the point-wise convolution, $Cat[.]$ represents the concatenate operation, $Sigmoid(.)$ is the Sigmoid function.
	
	 Moreover, the contribution of each stage $i(i=0,1,2,\dots,a)$ is distinct, because of the dissimilarity resolution and the different levels of information contained. Exactly as representative democracy, the votes depend on the number of citizens in the region. Moreover, emphasizing low-level features such as intensity, clutter \cite{bib41}, and shape \cite{bib18} improves performance significantly.
	
	Thus, distinct weights should be provided to all of the $\hat{\mathbf{D}}_i$. RB derives the weights from adjusting the number of edge channels $\hat{C}_i$ dynamically:
	\begin{equation}
		\label{10}
		\hat{C}_i=\lceil{t\cdot2^{i-1}}\rceil,t\in \mathbb{R}^+,
	\end{equation}
	where $t$ is a hyper-parameter to control the scale. Considering the calculation cost, $0<t<3$ is preferable.
	
	To sum up, the output of the representative block, i.e. the final output $\mathbf{O}$ of the ILNet can be formulated as:
	\begin{equation}
		\label{11}
		\mathbf{O}=Conv(Cat[\mathbf{F}_0, \mathbf{F}_1, \dots, \mathbf{F}_a]),
	\end{equation}
	where $Conv(.)$ is the point-wise convolution, $Cat[.]$ represents the concatenate operation.
	
	It is worth noting that, the Binary Cross Entropy (BCE) loss is calculated on each decoder and back propagate to support detection \cite{bib35}. This is equivalent to integrating models of distinct depths, which share weight partially.
	
	\begin{equation}
		BCE=\mathbf{GT}log(\mathbf{\hat{D}_i})+(1-\mathbf{GT})log(1-\mathbf{\hat{D}_i}),
	\end{equation}
	where $BCE$ denotes the BCE loss. $\mathbf{GT}$ is the ground truth.

	\begin{table*}[thbp]
		\centering
		\caption{Comparisons with SOTA methods on NUAA-SIRST and IRSTD-1k in $IoU(\%)$, $nIoU(\%)$, $P_d(\%)$ and $Fa(10^{-6})$. The best results are in \textcolor{red}{\textbf{red}}, the second results are in \textcolor{blue}{\textbf{blue}} and the third are in \textbf{bold black}.}
		\label{Li.t2}
		\renewcommand\arraystretch{1.5}
		\begin{tabular}{c|c|cccc|cccc}
			\Xhline{1.3pt}
			\multirow{2}{*}{Methods}&\multirow{2}{*}{Description}&\multicolumn{4}{c|}{NUAA-SIRST}&\multicolumn{4}{c}{IRSTD-1K}\\
			\cline{3-10}	&&$IoU$&$nIoU$&$P_d$&$F_a\downarrow$&$IoU$&$nIoU$&$P_d$&$F_a\downarrow$\\ 
			\Xhline{1pt}
			WSLCM \cite{bib7}&Local Contrast&1.158&0.549&77.95&5446&3.452&0.678&72.44&6619\\
			TLLCM \cite{bib8}&Local Contrast&1.029&0.905&79.09&5899&3.311&0.784&77.39&6738\\
			Max-Median \cite{bib5}&Filtering&4.172&2.15&69.20&55.33&6.998&3.051&65.21&59.73\\
			Top-Hat \cite{bib6}&Filtering&7.143&5.201&79.84&1012&10.06&7.438&75.11&1432\\
			MSLSTIPT \cite{bib12}&Local Rank&10.30&9.58&82.13&1131&11.43&5.932&79.03&1524\\
			RIPT \cite{bib10}&Local Rank&11.05&10.15&79.08&22.61&14.11&8.093&77.55&28.31\\
			NRAM \cite{bib9}&Local Rank&12.16&10.22&74.52&13.85&15.25&9.899&70.68&16.93\\
			PSTNN \cite{bib11}&Local Rank&22.40&22.35&77.95&29.11&24.57&17.93&71.99&35.26\\
			IPI \cite{bib13}&Local Rank&25.67&24.57&85.55&11.47&27.92&20.46&81.37&16.18\\
			\Xhline{1pt}
			MDvsFAcGAN \cite{bib16}&CNN&60.30&58.26&89.35&56.35&49.50&47.41&82.11&80.33\\
			AGPCNet \cite{bib20}&CNN&72.10&70.24&80.73&7.23&57.03&52.55&88.55&16.28\\
			ACM \cite{bib14}&CNN&71.57&72.77&98.15&34.47&58.43&54.34&89.23&23.57\\
			ALCNet \cite{bib15}&CNN&74.31&73.12&97.34&20.21&62.05&59.58&92.19&31.56\\
			UIUNet \cite{bib23}&CNN&75.39&74.67&97.25&42.41&66.64&60.20&89.23&16.02\\
			DNANet \cite{bib19}&CNN&77.47&75.82&98.48&12.86&64.29&63.47&95.29&19.78\\
			ISNet\footnotemark[1] \cite{bib18}&CNN&\textcolor{blue}{\textbf{80.02}}&\textcolor{blue}{\textbf{78.12}}&\textcolor{blue}{\textbf{99.18}}&\textbf{4.924}&\textcolor{blue}{\textbf{68.77}}&\textbf{64.84}&\textcolor{red}{\textbf{95.56}}&15.39\\
			\Xhline{1.3pt}
			\textbf{ILNet-S(ours)}&CNN&78.12&76.42&\textbf{99.07}&5.50&66.01&64.78&93.27&\textbf{5.26}\\				
			\textbf{ILNet-M(ours)}&CNN&\textbf{79.57}&\textbf{77.19}&98.15&\textcolor{blue}{\textbf{3.02}}&\textbf{67.86}&\textcolor{blue}{\textbf{68.40}}&\textbf{94.61}&\textcolor{blue}{\textbf{5.09}}\\				
			\textbf{ILNet-L(ours)}&CNN&\textcolor{red}{\textbf{80.31}}&\textcolor{red}{\textbf{78.22}}&\textcolor{red}{\textbf{100}}&\textcolor{red}{\textbf{1.33}}&\textcolor{red}{\textbf{70.15}}&\textcolor{red}{\textbf{68.91}}&\textcolor{blue}{\textbf{95.29}}&\textcolor{red}{\textbf{3.23}}\\
			\Xhline{1.3pt}
		\end{tabular} 
	\end{table*}

	\section{Experiment}
	In this section, we first introduce the datasets and evaluation metrics. Then, the training protocols are introduced. Next, we visualize and compare our ILNet with several SOTA methods. Finally, we present several ablation studies of our network.

	\subsection{Evaluation Metrics and Datasets}
	\textbf{Datasets}: Our experiments are conducted on the NUAA-SIRST dataset \cite{bib14} and the IRSTD-1k dataset \cite{bib18}. The NUAA-SIRST dataset has 341 training data and 86 testing data. The IRSTD-1k dataset has 800 training data and 201 testing data.
	
	\textbf{Evaluation Metrics}: \textit{1) Intersection over Union(IoU)}: $IoU$ IoU is the most common pixel-level evaluation metric of infrared small target detection, which is calculated over the entire dataset. It is the ratio of the intersection and union region between the predictions and the ground truth. The correctness of each pixel has a great impact.
	\begin{equation}
		\label{12}
		IoU=\frac{A_i}{A_u}=\frac{\sum\limits_{i=1}^{n}{TP}_i}{\sum\limits_{i=1}^{n}T_i+P_i-{TP}_i},
	\end{equation}
	where $A_i$ and $A_u$ are the intersection and union, respectively. $T$ denotes the pixels predicted as the targets. $P$ denotes the pixels of the ground truth targets. $TP$ is the true positive pixels. $n$ represent the number of infrared images in the test set.
	
	\textit{2) Normalized Intersection over Union (nIoU)}: $nIoU$ is the arithmetic mean of $IoU$ for each sample \cite{bib14}.
	\begin{equation}
		\label{13}
		nIoU=\frac{1}{n}\sum\limits_{i=1}^{n}\frac{{TP}_i}{T_i+P_i-{TP}_i}.
	\end{equation}

	\textit{3) Probability of Detection ($P_d$)}: $P_d$ evaluates the ratio of the true targets detected by the model to all the ground truth targets. It is a target-level evaluation metric. A detected target is considered as $TP$ if the centroid derivation is less than 3 \cite{bib19}.
	\begin{equation}
		\label{14}
		P_d=\frac{{TP}_{sum}}{T_{sum}},
	\end{equation}
	where ${TP}_{sum}$ represents the number of detected targets, $T_{sum}$ represents all the targets in the test set.
	
	\textit{4) False-Alarm Rate (Fa)}: A target-level evaluation metric, to measure the ratio of predicted target pixels that do not match the ground truth.
	\begin{equation}
		\label{15}
		F_a=\frac{\sum\limits_{i=1}^{n}{FP}_{i}}{ALL},
	\end{equation}
	where $FP$ denotes the number of false alarm pixels, and $ALL$ is all pixels in the image.

	\footnotetext[1]{The original data is used[20]. We cannot reproduce enough precision due to low computing power.}
	
	\subsection{Implementation Details}
	Adan \cite{bib42} is utilized as the optimizer of the ILNet and its hyper-parameters are set to default (initial learning rate lr=1e-3, betas=(0.98, 0.92, 0.99), eps=1e-8, \textit{max\_grad\_norm}=0), except \textit{weight\_decay}=1e-4. Learning rate decay is used to train our network. The input images are resized to 512×512 during training and testing. Binary cross entropy loss is used as the loss function. AMP(Automatic Mixed Precision) \cite{bib43} and learning rate decay is utilized to train 600 epochs, with a batch size of 8. Our ILNet is implemented based on Pytorch 1.10.0 \cite{bib44} and conducted on an NVIDIA Jetson AGX Xavier.
	
	Detailed configurations of the ILNet are presented in Table \ref{Li.t1}. Our ILNet-M follows the small size U$^2$Net \cite{bib35}, ILNet-L is a slightly larger channel-anti-bottleneck structure, while ILNet-S is a relatively minimal network. Different model architectures are applied to different application scenarios. ILNet-L is suitable for devices with sufficient computing resources, while the small one is more suitable for edge devices such as NVIDIA Jetson AGX Xavier.

	\begin{figure*}[!t]
		\centering
		\includegraphics[width=7.4in]{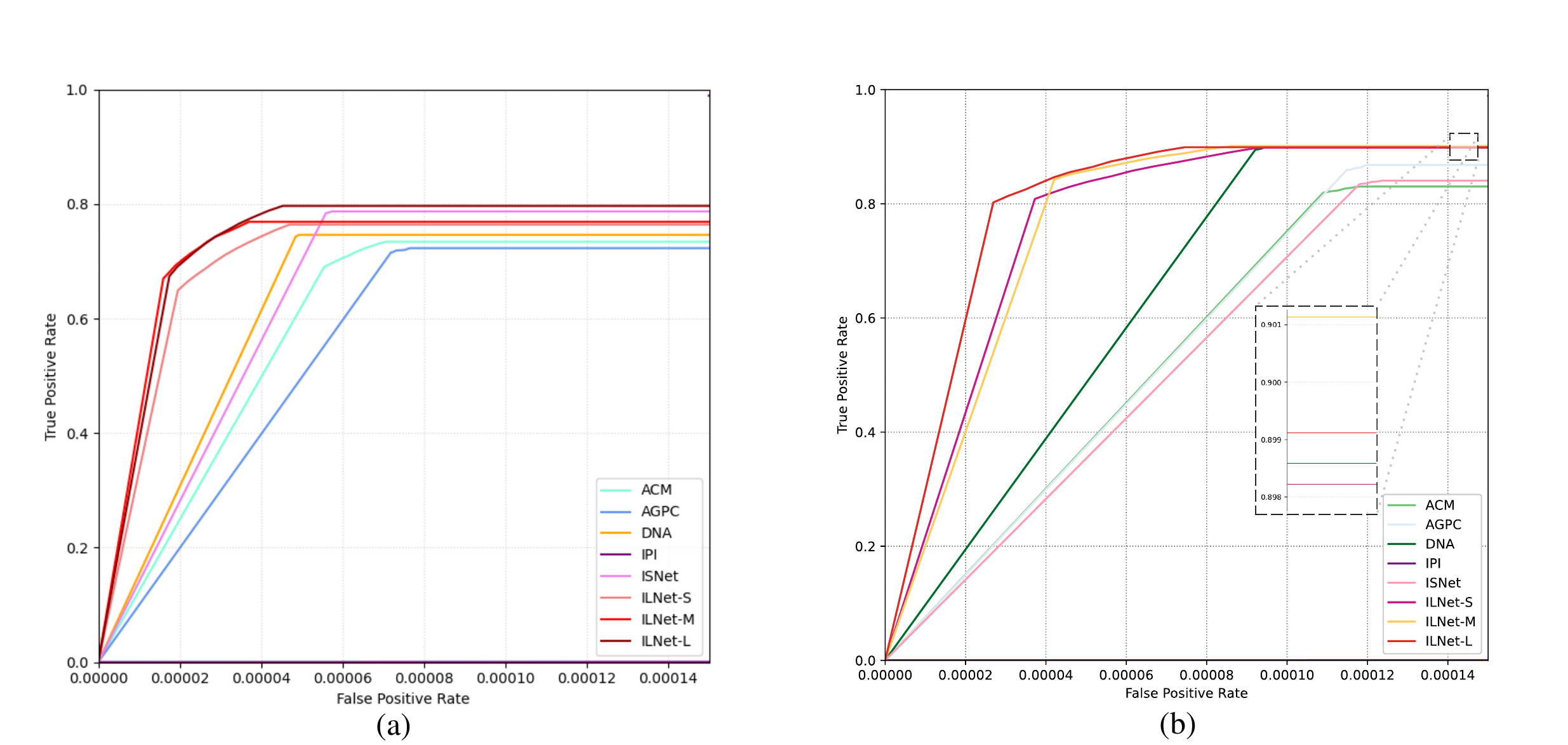}%
		\caption{ROC curves of different methods on the (a) NUAA-SIRST and (b) IRSTD-1K datasets.}
		\label{Fig9}
	\end{figure*}

	\begin{figure}[!t]
		\centering
		\includegraphics[width=3.8in]{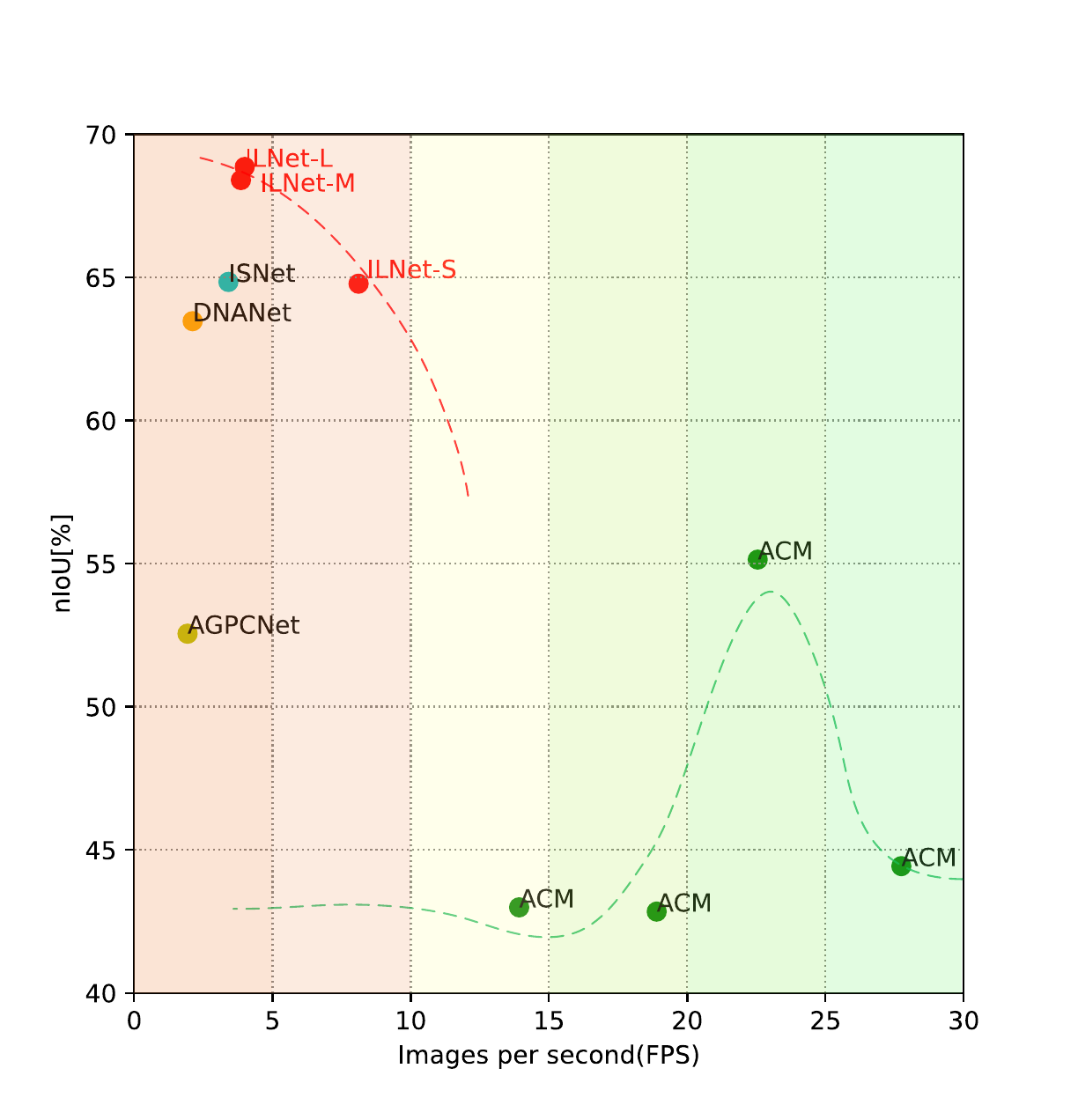}
		\caption{nIoU \emph{vs.} number of images per second (with batch size 1) using the NVIDIA Jetson AGX Xavier (without TensorRT acceleration). This is the average of 500 images.}
		\label{Fig10}
	\end{figure}

	\begin{figure*}[!t]
		\centering
		\setlength{\abovecaptionskip}{0.cm}
		\includegraphics[width=7.2in]{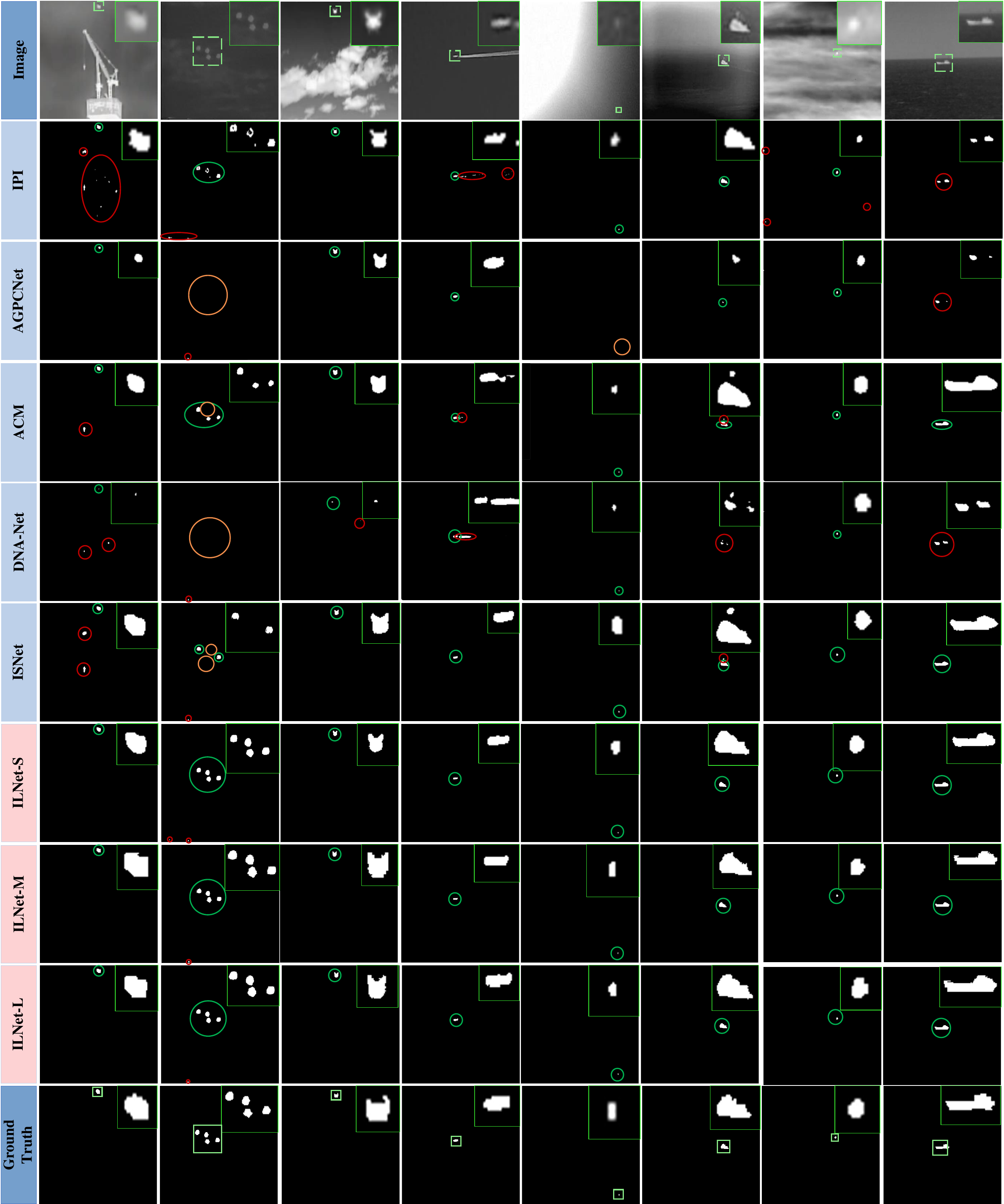}
		\caption{Qualitative results obtained by different methods on the NUAA-SIRST dataset. Enlarged targets are shown in the right-top corner. Circles in green, red and orange represent correctly detected targets, false alarm, and miss detected targets, respectively.}
		\label{Fig11}
		\vspace{-0.4cm}
	\end{figure*}

	\begin{figure*}[!t]
		\centering
		\setlength{\abovecaptionskip}{0.cm}
		\includegraphics[width=7.2in]{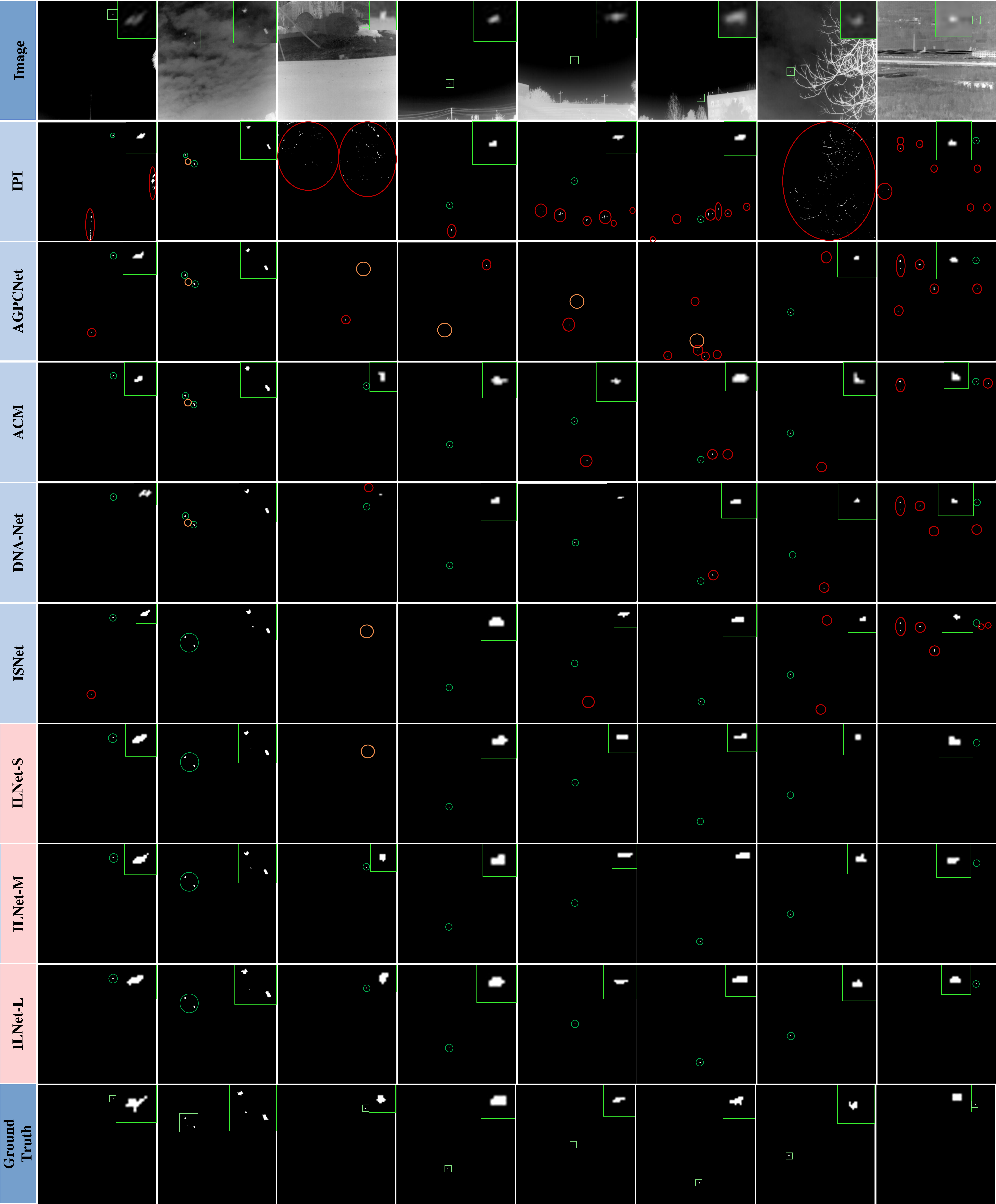}
		\caption{Qualitative results obtained by different methods on the IRSTD-1K dataset. Enlarged targets are shown in the right-top corner. Circles in green, red and orange represent correctly detected targets, false alarm, and miss detected targets, respectively.}
		\label{Fig12}
		\vspace{-0.4cm}
	\end{figure*}

	\subsection{Comparison to the State-of-the-art Methods}
	The ILNet is compared with 15 SOTA methods including traditional methods: WSLCM \cite{bib7}, TLLCM \cite{bib8}, Max-Median \cite{bib5}, Top-Hat \cite{bib6}, MSLSTIPT \cite{bib12}, RIPT \cite{bib10}, NRAM \cite{bib9}, PSTNN \cite{bib11}, IPI \cite{bib13}; and deep learning methods: MDvsFAcGAN \cite{bib16}, ACM \cite{bib14}, ALCNet \cite{bib15}, ISNet \cite{bib18}, DNANet \cite{bib19}, AGPCNet \cite{bib20}.
	
	\textit{1) Quantitative Results}: As shown in Table \ref{Li.t2}, ILNet has a significant improvement over other deep learning methods. Especially on the larger dataset IRSTD-1K, our ILNet-L achieves +1.38 IoU and +4.07 nIoU compared with the SOTA method ISNet \cite{bib18}. Meanwhile, our method also makes tremendous progress on the target-level evaluation metrics. The false alarm rates on two datasets decreased by 72.99\% and 92.20\%, respectively. Thus, as the dataset expands, our ILNet will be significantly improved. Astonishingly, experiments demonstrate that the smallest ILNet still performs superior, and offers a better trade-off between floating point operations (2.19GFLOPs), parameters (0.046M), and accuracy. Consequently, ILNet achieves accuracy at a more efficient computing cost, as shown in Fig.\ref{Fig1} and Table \ref{Li.t2}. It is considered that ILNet focuses more on low-level information, and incorporates crucial low-level features into deep layers. Moreover, by using shallow features to guide up-sampled deep features, small infrared targets in the deep network are enhanced. This improves the deep abatement to some extent.
	
	The ROC curves shown in Fig.\ref{Fig9} (a) and Fig.\ref{Fig9} (b) demonstrate the superiority and stability of ILNet.
	
	ILNet series better balance accuracy with the FPS (number of images per second), as shown in Fig.\ref{Fig10}. They can exchange accuracy for speed within a certain range, or speed for higher accuracy, to adapt to different needs. ACM \cite{bib14} is faster, but its potential for accuracy is limited.
	
	\textit{2) Qualitative Results}: Qualitative results are obtained by different methods on the NUAA-SIRST and IRSTD-1K datasets. Traditional methods (such as IPI) have a significant qualitative gap, therefore we mainly compare ILNet with deep learning methods. With more attention to low-level information, ILNet has a stronger edge-fitting ability and to separate targets and background at the pixel-level. ILNet has fewer miss detection and false alarm, as shown in Fig.\ref{Fig11} and Fig.\ref{Fig12}. ILNet relatively ignores high-level semantic information, thus is less prone to be in miscarriage of justice.
	
	Our study state clearly that on the complex IRSTD-1K datasets, most methods prefer to find relatively bright spots in the infrared images, due to the dataset propensity. This brings numerous false alarm. This propensity does not exist on our ILNet, as shown in Fig.\ref{Fig12}. 
	
	Fig.\ref{Fig8} indicates the improvement for deep targets recovery with the proposed representative block (RB). RB improves deep abatement to some extent through feature enhancement, which introduces low-level information into the deep stages of the network. Small targets are more precisely positioned and the edges are clearer with RB.
	
	\subsection{Ablation Study}
	In this subsection, we present three ablation experiments on NUAA-SIRST datasets, to verify the details and effectiveness of several blocks in our ILNet.
	
	\textit{1) Blocks Ablation}: The blocks ablation is to validate the effectiveness of the IPOF, DODA layers, and RB block we designed. As shown in Table \ref{Li.t3}, all of the blocks can improve the performance of the baseline, and the combination possesses a better effect. Among them, the IPOF w/o DODA has improved in IoU. With the addition of DODA, the performance of IPOF in the false alarm has been improved substantially. However, there are more miss targets. RB not only considerably improves the performance on pixel-level, but also significantly reduces the false alarm and miss detection. Even some variants can accurately detect all the targets in the NUAA-SIRST test set (the centroid derivation of the target is less than 3 pixels \cite{bib19}). It is an essential component for ILNet to accurately locate targets. Incorporating shallow features into the deep layers can alleviate target-level miss detection with IPOF only. RB further improves pixel-level performance.

	\begin{table}[!t]
		\centering
		\caption{Ablation study of the IPOF, DODA and RB block.}
		\label{Li.t3}
		\renewcommand\arraystretch{1.5}
		\begin{tabular}{c|cccc}
			\Xhline{1.3pt}
			Model&$IoU$&$nIoU$&$P_d$&$F_a\downarrow$\\ 
			\Xhline{1pt}
			ILNet w/o IPOF/RB&73.78&73.12&98.17&38.10\\
			ILNet+IPOF w/o DODA&75.72&74.23&97.25&25.02\\
			ILNet+IPOF&77.45&75.62&99.07&15.13\\
			ILNet+RB&77.21&74.76&\textbf{100}&7.76\\
			ILNet+IPOF+RB&\textbf{78.12}&\textbf{76.42}&99.07&\textbf{5.50}\\
			\Xhline{1.3pt}
		\end{tabular} 
	\end{table}

	\begin{table}[!t]
		\centering
		\caption{Ablation study on the different number of IPOF module.}
		\label{Li.t4}
		\renewcommand\arraystretch{1.5}
		\begin{tabular}{c|cccccc}
			\Xhline{1.3pt}
			\makecell[c]{Num of\\ Stages}&Params(M)&GFLOPs	&$IoU$&$nIoU$&$P_d$&$F_a\downarrow$\\ 
			\Xhline{1pt}
			0&0.036&1.29&73.78&73.12&98.17&38.10\\
			1&0.038&1.29&75.30&74.77&97.22&35.13\\
			2&0.039&1.30&76.37&75.56&99.07&26.97\\
			3&0.041&1.32&77.14&75.24&97.22&18.94\\
			4&0.042&1.42&77.37&75.55&98.15&28.21\\
			5&0.044&1.81&77.45&75.62&99.07&15.13\\
			\Xhline{1.3pt}
		\end{tabular} 
	\end{table}

	\begin{table}[!t]
		\centering
		\caption{Different modes of feature enhancement in RB block,i.e. RB+enh.H is using deep features to enhance shallow stages, RB+enh.L is opposite.}
		\label{Li.t5}
		\renewcommand\arraystretch{1.5}
		\begin{tabular}{c|cccc}
			\Xhline{1.3pt}
			Modes&$IoU$&$nIoU$&$P_d$&$F_a\downarrow$\\ 
			\Xhline{1pt}
			RB+enh.H&76.08&74.52&98.15&37.88\\
			RB+enh.L&\textbf{77.21}&\textbf{74.76}&\textbf{100}&\textbf{7.76}\\
			\Xhline{1.3pt}
		\end{tabular} 
	\end{table}

	\begin{table}[!t]
		\centering
		\caption{RB block with different up-sample methods. Enh.L is the proposed feature enhancement.}
		\label{Li.t6}
		\renewcommand\arraystretch{1.5}
		\begin{tabular}{c|cccc}
			\Xhline{1.3pt}
			\makecell[c]{Up-sample\\
				Methods}&$IoU$&$nIoU$&$P_d$&$F_a\downarrow$\\ 
			\Xhline{1pt}
			RB+nearest&75.44&74.16&99.08&30.16\\
			RB+bicubic&75.59&73.40&98.17&24.35\\
			RB+Carafe&76.19&74.12&98.17&8.87\\
			RB+bilinear&76.56&74.41&100&10.25\\
			RB+enh.L&\textbf{77.21}&\textbf{74.76}&\textbf{100}&\textbf{7.76}\\
			\Xhline{1.3pt}
		\end{tabular} 
	\end{table}
	
	\textit{2) Interactive Polarized Orthogonal Fusion Module (IPOF)}: This study ablates the design of the proposed IPOF module, by removing IPOF modules one by one from the shallowest stage, as shown in Table \ref{Li.t4}. There is no evident influence on the pixel-level, but for the target-level, more feature fusion will improve the performance. Considering the small GFLOPs and parameters, all the features of $\mathbf{E}_i(i=0,1,\dots, a)$ and $\mathbf{D}_i(i=0,1,\dots, a)$ are fused.
	
	\textit{3) Representative Block (RB)}: This study contrast the distinct feature enhancement modes for the deep and shallow stages, as shown in Fig.\ref{Fig2}. As shown in Table \ref{Li.t5}, the effects on pixel-level are not significant, but a significant impact exists in the target-level performance. Using shallow features to enhance the deep stages receives better performance.
	
	This study suggests that the traditional up-sampling methods do not restore the information of small infrared targets in the deep stages, as shown in Fig.\ref{Fig3}, which is the dominant reason behind deep abatement. We ablate the design of the proposed RB block by replacing various up-sampling methods, to verify which one is better to recover deep targets, as shown in Table \ref{Li.t6}. The proposed method is contrasted with traditional methods including Nearest-neighbor, Bilinear, Bicubic interpolation; and the deep learning method Carafe \cite{bib45}.The results demonstrate that the enhancement method we designed is admirably succinct and effective.
	
	\section{Conclusion}
	In this paper, we have presented a new network for infrared small target detection, ILNet, which considers infrared small targets as significant regions without much semantic information. An IPOF module for bidirectional feature fusion is designed, and adaptively aggregates channel and spatial information through DODA layers. In addition, an RB block is proposed to improve the deep abatement and fuse global features dynamically. Experimental results on two datasets demonstrate that ILNet achieves superior performance against other methods.
	
	However, since the existing datasets possess a small data quantity, the performance of the deep learning model is nearly saturated. The data problem seriously restricts the generalization and development of this research field, which should be a priority in the future.
	
	\bibliographystyle{IEEEtran}
	\bibliography{double-bibliography}
	
	\newpage
	
	\vspace{11pt}
	
	\begin{IEEEbiography}[{\includegraphics[width=1in,height=1.25in,clip,keepaspectratio]{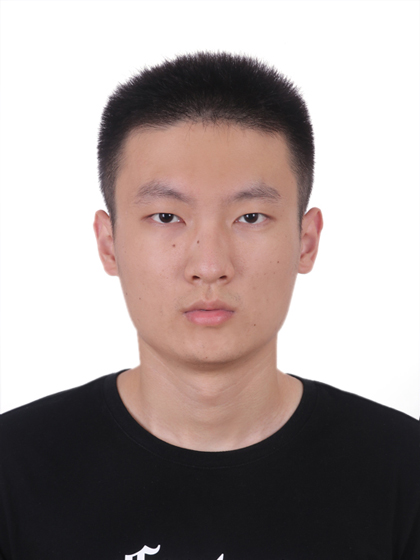}}]{Haoqing Li}
	received his B.Sc. degree in automation from Shandong University of Science and Technology, Qingdao, China, in 2021. He is currently pursuing an M.Sc. degree with the Department of Control Science and Engineering, Beijing University of Technology, Beijing. His research interests include computer vision, image semantic segmentation, deep learning, and their applications in infrared image processing.
	\end{IEEEbiography}

	\begin{IEEEbiography}[{\includegraphics[width=1in,height=1.25in,clip,keepaspectratio]{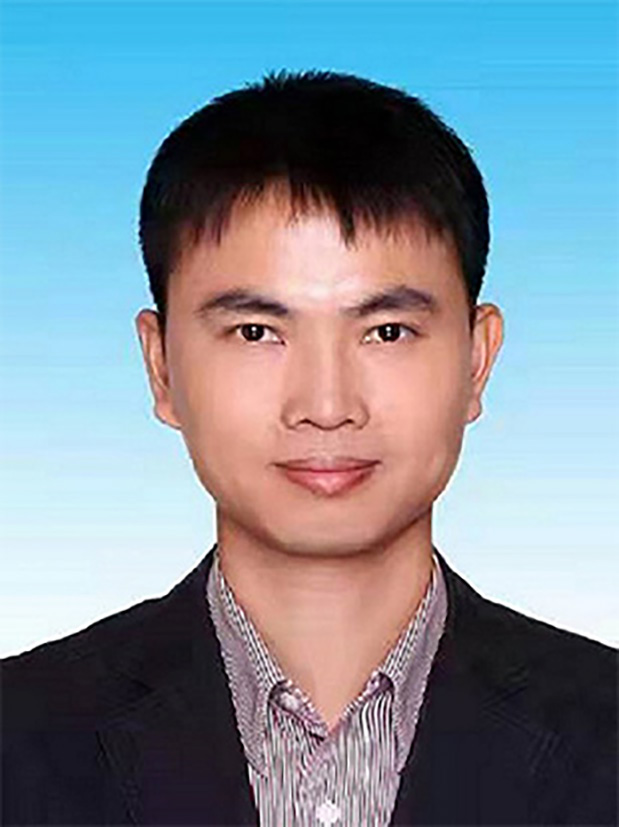}}]{Jinfu Yang}
	received his Ph.D. in pattern recognition and intelligent systems from the National Laboratory of Pattern Recognition, Chinese Academy of Sciences, in 2006. In 2013 - 2014, he was a visiting scholar at the University of Waterloo, Canada. Since 2006, he has been with the Faculty of Information Technology and the Beijing Key Laboratory of Computational Intelligence and Intelligent System, Beijing University of Technology, Beijing, where he is currently a Professor of computer vision. He is the secretary general of the Beijing Association for Artificial Intelligence, and a member of the Technical Committee of Computer Vision and Intelligent Robot of the Chinese Computer Federation (CCF CV and CCF TCIR). His research interests include pattern recognition, computer vision, and robot navigation.
	\end{IEEEbiography}

	\begin{IEEEbiography}[{\includegraphics[width=1in,height=1.25in,clip,keepaspectratio]{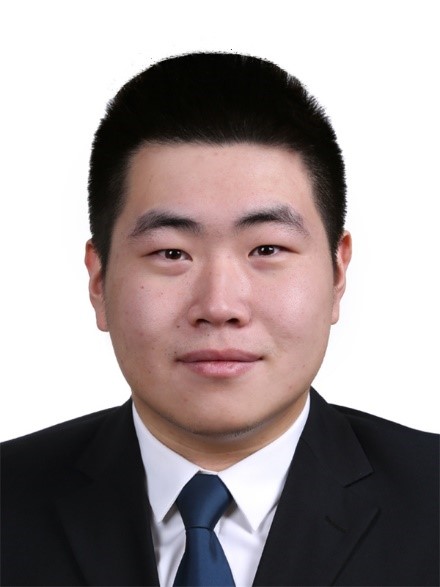}}]{Runshi Wang}
	 received his B.Sc. degree in automation from Beijing University of Technology, in 2021. He is currently pursuing an M.Sc. degree with the Department of Control Science and Engineering, Beijing University of Technology, Beijing. His current research interests include small target detection, image-to-image translation computer vision, and robot environment perception.
	\end{IEEEbiography}

	\begin{IEEEbiography}[{\includegraphics[width=1in,height=1.25in,clip,keepaspectratio]{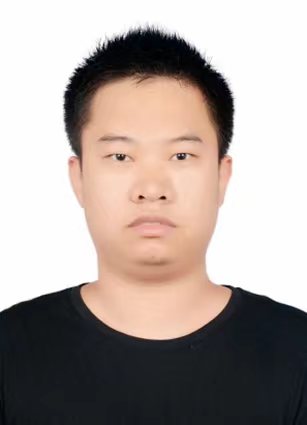}}]{Yifei Xu}
	received his B.Sc. degree in Measurement and control technology and instrument from Tianjin Polytechnic University, in 2021. He is currently pursuing an M.Sc. degree with the Department of Control Science and Engineering, Beijing University of Technology, Beijing. His research interests include deep neural network compression, computer vision, and robotic environmental awareness.
	\end{IEEEbiography}

	\vspace{11pt}
	
	\vfill
	
\end{document}